\documentclass{article} 
\usepackage{collas2026_conference,times}
\usepackage{easyReview}

\usepackage{amsmath,amsfonts,bm}

\def\eqref#1{equation~\ref{#1}}

\def\1{\bm{1}}

\DeclareMathAlphabet{\mathsfit}{\encodingdefault}{\sfdefault}{m}{sl}
\SetMathAlphabet{\mathsfit}{bold}{\encodingdefault}{\sfdefault}{bx}{n}

\usepackage{hyperref}
\hypersetup{
    colorlinks=true,
    linkcolor=red,
    filecolor=magenta,
    urlcolor=blue,
    citecolor=purple,
    pdftitle={Overleaf Example},
    pdfpagemode=FullScreen,
    }

\usepackage{booktabs} 
\usepackage{multirow}
\usepackage{algorithm}
\usepackage{algorithmic}
\usepackage{subcaption}

\title{Open-ended Multi-agent Autocurricula via Visual Inspection of Policies with Multi-modal LLMs}

\author{Lorenzo Pantè\textsuperscript{1} \quad Andrea Fanti\textsuperscript{1} \quad Roberto Capobianco\textsuperscript{2} \\
\textsuperscript{1}Sapienza University of Rome, Italy \quad \textsuperscript{2}Sony AI, Zurich, Switzerland \\
\texttt{pante.1885460@studenti.uniroma1.it} \quad \texttt{fanti@diag.uniroma1.it} \\ \texttt{roberto.capobianco@sony.com} \\
}

\preprintcopy 

\begin{document}

\maketitle

\begin{abstract}
Open-ended curricula in Reinforcement Learning (RL) aim to train generally-capable agents by identifying tasks that facilitate learning increasingly complex skills. A major challenge when designing such curricula is assessing task difficulty relative to the agent's current learning progress. While previous work has explored using scalar task scores or textual summaries of the agent's behavior, here we study a different approach: directly inspecting policy behavior via recorded episode videos. We introduce a simple yet effective instantiation of this approach which leverages a Video Language Model (VLM) to both process these videos and provide curriculum recommendations, which we call Visual Inspection of Policies (VIP). Since videos can naturally contain any number of controllable agents, we empirically study VIP on the StarCraft Multi-Agent Challenge (SMAC). We show that even with a lightweight and openly accessible VLM (VideoLLaMa2-7B), VIP can use policy videos to generate more effective curricula than both its text-only ablation and methods that rely on scalar task scores.
\end{abstract}

\begin{figure}[h]
    \begin{center}
        \includegraphics[width=0.75\textwidth]{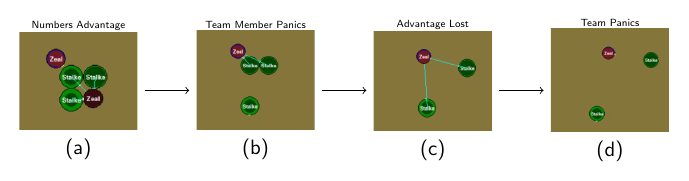}
        \caption{An example of how policy videos can capture promising curriculum directions that are inaccessible to methods based on learning signal. The four images are selected frames from an episode video recorded in the StarCraft Multi-Agent Challenge (SMAC) environment. While allied troops (green) have obtained a clear numbers advantage with an effective strategy in (a), they still end up losing this match to a single enemy (red) after panicking in (b) and (d). Despite the 0\% win rate, agents are close to discovering a winning strategy --- notably, this insight is inaccessible to any curriculum curator that relies on local learning signals (e.g.~Unsupervised Environment Design \citep{NEURIPS2020_985e9a46}). Here we instead propose a method to harness this information with a Vision Language Model (VLM), which we call \emph{Visual Inspection of Policies}. In this exact scenario, VIP chooses to continue training on the same task, achieving a substantial improvement to an $\sim$80\% win rate. Quantitative results supporting this intuition are reported in Section \ref{sec:experiments} (text-only ablation); other related qualitative analyses are also reported in Appendix \ref{sec:qualitative}.
        }
        \label{fig:demo}
    \end{center}
\end{figure}

\section{Introduction}

Many real-world Reinforcement Learning (RL) tasks are difficult or impossible to solve by simply starting from a random policy and directly trying to optimize its downstream performance \citep{10.1145/3321707.3321799,pmlr-v119-wang20l}. These tasks typically require intermediate ``stepping stone'' skills that are hard to discover with direct optimization and are impossible to identify in advance without extensive domain knowledge. This has motivated research into methods to automatically design \emph{open-ended} curricula \citep{Wang_Yang_Liu_Hao_Hao_Hu_Chen_Fan_Gao_2020,NEURIPS2020_985e9a46,ICLR2024_18b40f12,wang2023voyager} which both discover and solve increasingly complex tasks. The ultimate goal of these methods is to produce generally-capable agents that can quickly adapt to any task in a particular domain, including those unseen in training. To this end, the curriculum must recommend tasks that are neither too easy nor too hard, so that the agent is presented with the most interesting task that is still solvable. This tradeoff is typically highly dependent on the current learning progress of the agent: tasks that are easy and uninteresting to a very competent agent might still be worth tackling for an agent which has just started learning.

To correctly identify tasks that satisfy these properties, previous work has explored both scalar numerical scores \citep{pmlr-v139-jiang21b,10.5555/3737916.3738428, NEURIPS2021_0e915db6} and textual summaries of the agent's behavior \citep{ICLR2024_18b40f12,10626609}. The former typically estimate the learning signal that a task could provide, which can be limiting when the reward signal is sparse or deceptive. Many multi-agent games, for example, provide both a sparse reward signal (win rate) and heuristic dense rewards (e.g.~eliminating opponents) to aid exploration. However, since these heuristic rewards are not always correlated with winning strategies, relying solely on the learning signal can fail to recognize policies that are close to discovering successful strategies, as shown in Figure \ref{fig:demo}. Determining learning potential with text-only summaries \citep{ICLR2024_18b40f12,wang2023voyager} is limited for the same reason. Moreover, these approaches typically build on the human-like priors of LLMs to choose \emph{interesting} tasks; however, a human in the same position would typically rely on visual cues by observing the agent's behavior directly. A sports coach, for example, would much rather review the video of a match rather than only read descriptions of an athlete's performance. Nonetheless, this more direct visual approach has been inaccessible to RL autocurricula until the recent breakthrough of multi-modal LLMs and, more specifically, VLMs.

\begin{figure}
  \centering
  \includegraphics[width=1\textwidth]{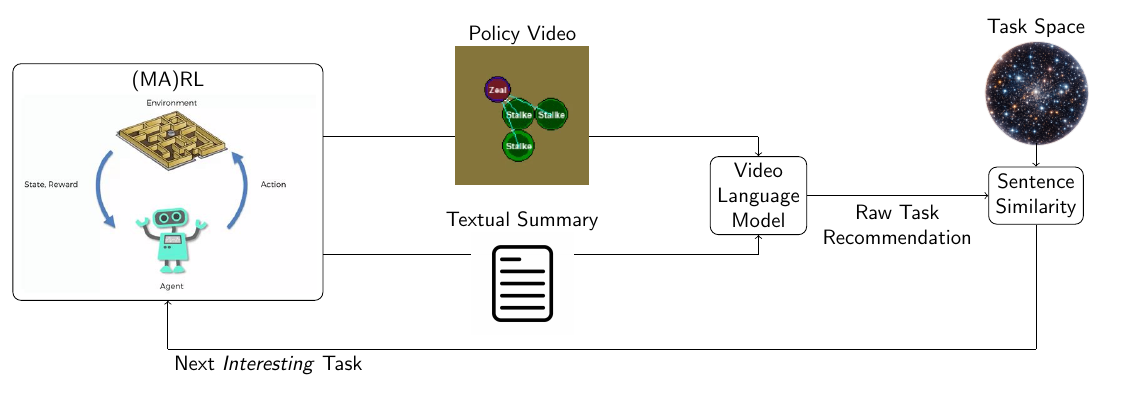}
  \caption{An overview of Visual Inspection of Policies (VIP). At each step of the curriculum, the agent is trained on the current task with a (Multi-Agent) Reinforcement Learning algorithm. Then, one or more recordings of episodes from the current policy are fed to the VLM, alongside a small textual summary of the agent's performance (e.g.~win rate). The VLM is prompted to recommend the next \emph{interesting} task, based on its assessment of the current capabilities of the agent. Finally, the raw recommendation from the VLM is compared with the set of tasks available to the curriculum (task space) using a sentence similarity module, to make sure the task is a valid specification.}
  \label{fig:arch}
\end{figure}

\paragraph{Contributions}
We bridge the gap discussed above by introducing a new approach to open-ended autocurricula: \emph{Visual Inspection of Policies} (VIP), outlined in Figure \ref{fig:arch}. VIP directly feeds episode videos of the current policy to a Vision Language Model (VLM) to gain direct access to visual information that might be missing in numerical or textual summaries. Additionally, VIP is independent of the number of agents in the environment as long as they are rendered in the same video. We apply VIP to the popular StarCraft Multi-Agent Challenge (SMAC), a complex, strategic, multi-agent collaborative domain with varying number and types of agents, previously unexplored in \emph{unsupervised} RL autocurricula. We find that VIP outperforms other autocurricula based on learning signal, and that episode videos play a crucial role in its effectiveness.

\section{Related work}
\paragraph{Automatic Curriculum Learning}
Open-ended autocurricula in Reinforcement Learning have their roots in Automatic Curriculum Learning (ACL). Standard ACL methods are supervised: they target a single final task and design a curriculum tailored to that task. \citet{Wang_Yang_Liu_Hao_Hao_Hu_Chen_Fan_Gao_2020} generates curricula in multi-agent environments by progressively incrementing the number of agents. PORTAL \citep{PORTAL} is another multi-agent ACL method which constructs the curriculum by evaluating the similarity between intermediate tasks. STAR-MARL \citep{li2026star} adopts an LLM to automatically generate appropriate sub-task trees from a known downstream MARL task.

\paragraph{Open-Ended Autocurricula}
In contrast to supervised ACL, open-ended autocurricula are fully unsupervised and do not target a single final task, assuming no knowledge on the identity, number or other properties of the downstream tasks. Open-ended autocurricula explicitly aim to continually propose worthwhile tasks to the learning agent and are closely related to the adjacent fields of lifelong learning and Continual Reinforcement Learning \citep{NEURIPS2023_9d8cf124}. A popular way to automatically design open-ended curricula is leveraging evolutionary computation methods, such as in POET and enhanced-POET \citep{10.1145/3321707.3321799,pmlr-v119-wang20l}: these co-evolve populations of agents and environments in a divergent process to discover and solve novel problems indefinitely. While highly influential, these works do not provide a clear mechanism for reusing the knowledge acquired by the agent population on unseen tasks, require substantial computational resources, and cannot be easily extended to multi-agent settings.

\paragraph{Unsupervised Environment Design}
Unsupervised Environment Design (UED) \citep{NEURIPS2020_985e9a46} is a more recent framework for open-ended curriculum generation based on minimax regret optimization. UED has been implemented both in teacher-student algorithms \citep{NEURIPS2020_985e9a46,NEURIPS2021_0e915db6} and in autocurricula based on estimates of learning potential \citep{pmlr-v139-jiang21b,NEURIPS2021_0e915db6,pmlr-v162-parker-holder22a}. Recent work has studied how these estimates might be vulnerable to selecting tasks that are too hard \citep{10.5555/3737916.3738428}, suggesting an alternative for boolean-outcome deterministic tasks. While these methods are designed for single-agent domains, UED has also been applied to multi-agent settings with MAESTRO \citep{samvelyan2023maestro}. However, MAESTRO is designed for two-player, adversarial, zero-sum games, and is not applicable to domains with higher or variable number of agents.

\paragraph{LLM-Guided Autocurricula}
The breakthrough of LLMs provided a new means to automatically design open-ended curricula, both to evaluate agent progress and propose new tasks. LLM-POET \citep{10626609} combines LLMs with the POET algorithm discussed above. OMNI and OMNI-EPIC \citep{ICLR2024_18b40f12,ICLR2025_d40d7cbe} employ the LLM as model of interest (MOI) biased towards tasks at the frontier of the agent capabilities that would also be interesting to a human. OMNI-EPIC also experiments with Video LMs but finds them to not be sufficiently reliable. However, their use of Video LMs is relegated to the role of success detectors, again falling into the bucket of compressing rich behaviors into brief summaries. Moreover, OMNI-EPIC relies on the domain exposing a way to program environments in code, which is realistically verified only for simulations that feature such functionality. For example, OMNI-EPIC cannot be applied to any simulation benchmark where physics and/or world dynamics are not properly decoupled from environment variants.
\citet{dharna2025foundation} propose a method that combines the power of the human priors contained in Foundation Models (FMs) with self-play. This method, however, scales with the number of agents and is not trivially extensible to variable or
high number of agents. \citet{zhang2024gameonlanguagemodels} propose a method with some similarities to VIP, which uses a VLM to guide open-ended RL training experiments, including recommending intermediate tasks in a curriculum. However, this method relies on manually identifying a set of skills before the start of the learning process, requiring significant prior domain knowledge and manual design.

\paragraph{Positioning}
All of the approaches discussed above use aggregate information on the current learning progress to choose the next best task. Scalar task scoring such as in UED might fail to capture nuanced information that is not directly linked to learning signals --- on the other hand, LLM-based methods mostly focus on describing the candidate tasks, rather than describing the learning progress of the current policy. In this work, we instead take advantage of VLMs to directly gauge the learning progress of the agents by analyzing episode videos of the current policy.

\section{Background}

\paragraph{Open-ended Reinforcement Learning}
Reinforcement Learning (RL) \citep{Sutton&Barto} deals with learning solutions to sequential decision problems, usually modeled via Markov Decision Processes (MDPs) or Partially-Observable Markov Decision Processes (POMDPs).
While (PO)MDPs can model a wide range of tasks, they assume that all components are stationary throughout the interaction with the agent. This fails to model many realistic scenarios which feature variations of one or more components after training, requiring the agent to generalize to unseen situations \citep{Fran_ois_Lavet_2018}. These settings can instead be modeled with a distribution $\Delta(\Theta)$ over a set $\Theta$ of (PO)MDPs. Recent work has then started to consider cases where only the set $\Theta$ of possible (PO)MDPs is known at training time, while the target distribution $\Delta(\Theta)$ is either unknown or absent, leading to \emph{open-ended} RL \citep{NEURIPS2020_985e9a46,DBLP:journals/corr/abs-2312-03126, pmlr-v119-cobbe20a, baker2020emergenttoolusemultiagent, hafner2024masteringdiversedomainsworld, pmlr-v202-bauer23a}. In open-ended RL, agents do not have a predefined \emph{final} objective and must adapt promptly to any possible situation in the given domain. One of the most prominent approaches to open-ended RL is based on Automatic Curriculum Learning (ACL) \citep{ijcai2020p671}, where agents are first trained on a sequence of training tasks, and then transferred to a set of test tasks \citep{NEURIPS2020_985e9a46, baker2020emergenttoolusemultiagent}, usually composed of tasks left out during training, to assess the generalization power of the trained agent. The selection of each task is, in general, based on the current learning progress of the agent, i.e. its current decision-making policy, and all previously selected tasks. In practice, however, existing approaches avoid directly examining the policy, and instead perform task selection from high-level aggregated information, such as performance \citep{10.1145/3321707.3321799,10626609}, learning potential \citep{NEURIPS2020_985e9a46}, novelty \citep{10.1145/3321707.3321799}, and/or human-like notions of interestingness \citep{ICLR2024_18b40f12}.

\paragraph{Multi-Agent Reinforcement Learning}
VIP can in principle be applied to any type of single or multi-agent problem. Here we summarize the formulation of multi-agent collaborative decision problems, which is the type of domain used in our experiments. We then model each task in our curriculum as a \emph{Decentralized} POMDP defined as a tuple $(N, S, A, T, R, \gamma)$, where: $N$ is the number of players (agents); $S$ is the environment state space; $A$ is the \emph{joint} action space $A = A_{1} \times \dots \times A_{N}$, with $A_i$ being the action space of agent $i$; ${T}: S \times A \to \Delta(S)$ is the environment transition function; $R: S \times A \times S \to [0,1]$ is the team reward function; $\gamma \in [0,1)$ is a discount factor. The objective in each single task is to maximize the discounted cumulative reward $G = \sum_{t=0}^\infty \gamma^t R_t$.

\paragraph{Large Language Models}
Large Language Models (LLMs) have emerged as a significant advancement in the field of Natural Language Processing \citep{minaee2024largelanguagemodelssurvey, 10.1145/3744746}. LLMs are transformers \citep{NIPS2017_3f5ee243} with a large number of learning parameters.
Although LLMs can be discriminative or generative models, it is common to refer to the latter as simply LLMs, which we also do here. Large Language Models demonstrated consistently superior performances across several tasks when compared to previous techniques \citep{NEURIPS2020_1457c0d6}. On the other hand, LLMs require significantly more computational power \citep{Samsi2023FromWT, hu2024optimizinglargelanguagemodels} due to the transformer architecture \citep{ansar2024surveytransformersnlpfocus, visiope_paper}. This can pose severe practical limitations to their use both in industry and research. Although LLMs were initially developed to only process and output text, the recent breakthrough of Multi-modal LLMs \citep{caffagni-etal-2024-revolution} has allowed them to process multi-modal data, such as images and videos. In this work we specifically use Video Language Models (VLMs): multi-modal LLMs that accept text and videos as input modalities and output text.

\section{Visual Inspection of Policies}
\label{sec:method}

\begin{algorithm}[t]
\caption{Visual Inspection of Policies (VIP)}
\label{alg:algorithm}
\textbf{Input}: Initial agents $A^i_0$ with policies $\pi^i_0$, initial task $\theta_0$. \\
\textbf{Parameter}: Curriculum step $t$, task $\theta_t$, agents $A^i_t$ 
with policies $\pi^i_t$, policy videos $\phi_t^{\text{video}}$, text summaries $\phi_t^{\text{text}}$, VLM output $\vartheta_t$. \\
\textbf{Output}: Agents $A^T_i$ trained on the curriculum $\theta_0 \dots \theta_T$.
\begin{algorithmic}[1] 
\STATE $t \gets 0$.
\WHILE{True}
    \STATE $\pi^i_t \gets$ $\text{Train-RL}(A^i_t, \theta_t)$.
    \STATE $\phi_t^{\text{video}}, \phi_t^{\text{text}} \gets \text{Eval}(\pi^i_t, \theta_t)$.
    \STATE $\vartheta_t \gets \text{VLM-Next-Task}(\phi_t^{\text{video}}, \phi_t^{\text{text}})$.
    \STATE $\theta_t \gets \text{Sanitize-Task}(\vartheta_t)$.
    \STATE $t \gets t + 1$.
\ENDWHILE
\end{algorithmic}
\end{algorithm}

Here we present Visual Inspection of Policies (VIP), a new method to generate curricula using a Vision Language Model to inspect one or more episode videos of the current policies. VIP employs a single VLM to both analyze episode videos and recommend appropriate \emph{interesting} tasks for the curriculum based on its human-like priors. As depicted intuitively in Figure \ref{fig:demo}, the added visual information allows VIP to take into account learning progress cues that might be missing from other behavior summarization modalities. VIP's intentional simplicity makes it easy to test this intuition empirically with a text-only ablation while still generating effective curricula (Section \ref{sec:experiments}). An overview of VIP is presented in Algorithm \ref{alg:algorithm}. First, policies are trained on the most recently recommended task with a Reinforcement Learning algorithm (Train-RL). The resulting policies are then evaluated on the current task to gather one or more videos and other information. This multi-modal information is then fed to the VLM, which is prompted to select the task that best facilitates learning (VLM-Next-Task). To mitigate hallucinations and formatting errors, the raw task recommendation of the VLM is then passed through a sentence-similarity module (Sanitize-Task) to obtain a valid task specification. The initial task in the curriculum can be either directly queried from the VLM or manually set where sensible, while policies are initialized according to the low-level RL algorithm (Train-RL), typically by randomly sampling model parameter values.

In principle, VIP is agnostic to the number of agents in the environment, as there is no hard constraint on the number of agents that may appear in a single video. In a multiplayer team game, for example, the VLM can directly process the video performances of the whole team, spending the same compute required to analyze a single policy. More generally, VIP requires prompting the VLM for $N$ task recommendations, where $N$ is the number of distinct models being trained --- any environment with a variable number of agents sharing a single model would then result in $N=1$. Another benefit of VIP's simplicity is that it introduces no new hyper-parameters in addition to those needed by the VLM and low-level RL algorithm. This is especially desirable for open-ended autocurricula since one of its overarching goals is precisely replacing brute-force grid search and similar supervised ways to acquire domain knowledge. We discuss this more in depth at the start of Section \ref{sec:experiments} and in Section \ref{grid search}. VIP also makes the basic assumption that the environment provides at least partial image observations or any kind of rendering functionality, which is trivially verified in most simulated and real-world RL domains. Finally, the effectiveness of VIP largely relies on the VLM, as with any LLM-based autocurricula \citep{ICLR2024_18b40f12}. By design, such methods cannot be truly robust to the choice of LLM/VLM, as they must naturally inherit specific weaknesses of the model to gain access to its strengths. At the same time, the LLM should be chosen such that it does not become a bottleneck in the curriculum. For this reason, our experiments employ a lightweight, openly accessible VLM which results in task recommendations only taking $\sim$1\% of the total curriculum training time (Appendix \ref{compute resources}).




\subsection{Prompting and Sanitization}
In order to produce sensible task suggestions, the VLM is provided with a \emph{system prompt} before the start of the curriculum. This system prompt should contain minimal information about the domain, the set of possible tasks, the syntax and semantics of the chosen task representation, and the goal of the VLM. The latter, in particular, can lead to curricula with different trade-offs between choosing novel tasks and tasks that are easy to learn for the current agents. At each step of the curriculum, the VLM is then fed a \emph{task prompt}, which should contain one or more video recordings of the current policies. Other numerical or textual information can also be included in the prompt, such as win rate or other performance metrics, estimates of learning potential or novelty, etc. In this work we only add the win rate and a list of the previously selected tasks to the videos. The latter turned out to be necessary in our experiments due to the context capabilities of the chosen VLM (see Appendix \ref{sec:prompt-templates}).

Even if provided with a precise description of the desired task specification format, there is no guarantee that the VLM will follow it in all outputs. Moreover, the VLM might hallucinate task specifications, especially when the syntax allows to express more tasks than those actually available. For example, the domain we chose for our experiments uses task specifications with letters and numbers, but not all combinations correspond to valid configurations. A basic solution, which we adopt in this work, is to use a simple module that outputs the best-matching valid task description from a VLM answer based on sentence similarity. This module computes the similarity between the sinusoidal embeddings \citep{NIPS2017_3f5ee243} of the VLM output and the candidate task names, returning the candidate with the highest similarity score. In our experiments, we also set a threshold such that when no task has sufficient similarity to the available ones, a random one is chosen, although this fallback mechanism was never triggered. Since the similarity module was able to recover all of the VLM failure modes in our experiments (Appendix \ref{sec:xp:sanitize}), we have refrained from employing alternatives such as iterative prompting \citep{wang2023voyager}, as these would have introduced unnecessary risks and additional VLM queries. The similarity module may also fail to recover hallucinations when the task specifications are heavily out-of-distribution w.r.t. the embedding model. In practice, this is rarely the case since most task specifications resemble natural language or human-readable configurations. Iterative prompting or training a suitable embedding model are both valid alternatives for edge cases with more obscure task specification formats.
This simple solution also does not immediately apply to non-enumerable task spaces --- these can also easily be handled with iterative prompting. Moreover, many non-enumerable task spaces adopt task specifications such that all syntactically correct specifications are valid task and thus require no sanitization.

\section{Experiments}
\label{sec:experiments}
We investigate the following research questions:
\begin{itemize}
    \item \textbf{Q1}: Does adding visual behavioral information (policy videos) to textual summaries result in more effective curricula?
    \item \textbf{Q2}: Can VIP outperform unsupervised autocurricula methods based on scalar task scores?
\end{itemize}

We study these questions on the StarCraft Multi-Agent Challenge (SMAC) \citep{samvelyan2019starcraftmultiagentchallenge}. SMAC is a cooperative, strategic multi-player game that tests individual and collaborative agent capabilities at several time scales, making it a popular benchmark for Multi-Agent Reinforcement Learning. The high variety in agent composition and environment configuration, however, makes SMAC particularly challenging for transfer and/or curriculum learning --- to the best of our knowledge, no previous work has tested \emph{unsupervised} autocurricula on this benchmark. We answer Q1 and Q2 by comparing VIP against: (i) its text-only ablation; (ii) two popular variants of Robust Prioritized Level Replay \citep{NEURIPS2021_0e915db6}; (iii) control baselines such as random curricula and training from scratch. All comparisons are performed both on held-out tasks (Sections \ref{test tasks}) and on tasks commonly seen by many curricula (Section \ref{seen tasks}).

In all curricula, agents are trained for $10^7$ environment steps on each task using Multi-Agent Proximal Policy Optimization (MAPPO) \citep{NEURIPS2022_9c1535a0}, a popular MARL algorithm. Since zero-shot transfer between configurations is unlikely to succeed in SMAC \citep{PORTAL}, we instead test generalization of policies produced by each curriculum by evaluating their win rate after fine-tuning on each test task \citep{baker2020emergenttoolusemultiagent,10.1145/3321707.3321799} for another $10^7$ steps. While this is a well-established protocol for SMAC \citep{NEURIPS2022_9c1535a0,PORTAL}, it introduces the issue of tuning hyper-parameters when applied to \emph{unsupervised} curricula. In fact, by definition, these cannot assume knowledge of the downstream task and thus cannot collect environment interactions for the purpose of grid search or similar methods. Moreover, optimal hyper-parameter values in SMAC are not reusable for multiple tasks as they often differ widely \citep{NEURIPS2022_9c1535a0}. Our main experiments in Sections \ref{test tasks} and \ref{seen tasks} share the same default MAPPO
hyper-parameters among all considered curricula (Table \ref{tab:hyper-parameters}), with the objective of avoiding reliance on downstream task knowledge. We also conduct a secondary set of experiments that compare VIP with \emph{supervised} grid search and discuss the results in Section \ref{grid search}. All fine-tuning curves and final performances are reported with 95\% confidence intervals over 5 independent seeds.

\subsection{Setup}

\paragraph{Environment}
The environment of SMAC consists of two opposing teams of troops, ``allied'' against ``enemies'', each shooting at the opposing faction within a 2D continuous plane, with the objective of eliminating all opponents. Learning agents correspond to the allied troops, while enemy troops are controlled by the game AI. Troops may vary in type and number across both teams. More details on SMAC as a set of (PO)MDPs, such as observation and action spaces, are contained in Appendix \ref{SMAC details}. For the purposes of computing the win rate, a round is won when all enemy troops are dead; all other cases are considered losses, including draws. To ensure fairness with respect to prior domain knowledge assumptions, we record episode videos with the built-in rendering scheme of SMAC, although it can sometimes be limited in accurately depicting the state of the game (see Appendix \ref{SMAC details}).

\paragraph{Task Space}
Tasks in SMAC are identified by two configuration parameters: the \emph{map}, which controls allied and enemy troop compositions, and the \emph{difficulty}, which controls the behavior of the game AI. Troops of different types have substantially different capabilities and thus require completely different strategies to be piloted successfully. We follow the map naming convention of \citet{vinyals2017starcraftiinewchallenge} --- for example, the map name \texttt{3s5z\_vs\_3s6z} specifies allies having 3 troops of type \texttt{s} (Stalkers) and 5 troops of type \texttt{z} (Zealot), and enemies having 3 Stalkers and 6 zealots. Maps where teams have the same composition only report the first portion, e.g.~\texttt{3s5z} is equivalent to \texttt{3s5z\_vs\_3s5z}. The difficulty is a simple integer in the range 1-10. Counterintuitively, there are some instances of maps being harder to solve at specific lower difficulties (see Appendix \ref{difficulty5} for examples). This is likely due to the game AI levels being tailored to human players. Note that even though infinite allied and enemy compositions exist, not all of them are valid tasks in SMAC. This means that the actual task space is a finite and enumerable subset of the full space of all conceivable compositions. Note that an enumerable task space embedded in an infinite task space is not necessarily easier for curriculum design, as it might miss better tasks that are outside the subset. At the same time, the full task space might contain more distractors, which in turn means that neither configuration can be definitively said to be easier or harder. To properly evaluate generalization, we split the available maps into an 80\% available for training (12 maps) and a 20\% for testing fine-tuning on unseen maps (3 maps). The subdivision is reported in Table \ref{tab:train test maps} and further motivated in Section \ref{test tasks}.

\paragraph{VLM}
We use VideoLLaMa2-7B, the 7-billion quantized version of the openly accessible VideoLLaMa2 \citep{Cheng2024VideoLLaMA2A}, as the VLM for VIP, paired with the sentence similarity model from \citet{sentencesimilarity}. Since \mbox{VideoLLaMa2} does not support video batching, each prompt is paired with one video of an episode sampled from the current team policy on the last selected task in the curriculum. Videos are fed as-is to the VLM without any pre-processing.
Despite these limitations, our experiments show that this VLM is sufficient for VIP to outperform the considered baselines. This choice also ensures that the VLM is not a bottleneck in the curriculum: in all our runs, MAPPO training strongly dominated raw compute times by taking up around 99\% of the time spent for each VIP run, whereas VLM inference was limited to the remaining 1\% (see Appendix \ref{compute resources}).
The textual summary fed to the VLM alongside the video consists simply of the win rate and a list of previously selected tasks. More details on VLM hyper-parameters, prompt templates, example VLM outputs and other VLM-related details can be found in Appendix \ref{vlm details}.

\paragraph{Baselines}
\label{sec:baselines}
As mentioned above, we answer Q1 by comparing VIP against a text-only ablation. The only difference between VIP and this ablation is that the video input and phrases mentioning it are omitted from the prompts, allowing to measure the contribution of episode videos to VIP's effectiveness. We also compare VIP against Robust Prioritized Level Replay (PLR$^\perp$) \citep{NEURIPS2021_0e915db6}, an Unsupervised Environment Design method \citep{NEURIPS2020_985e9a46} that selects tasks according to a scalar score function evaluated from policy rollouts. Here we compare to PLR$^\perp$ with two common choices for the score function: Positive Value Loss (PVL) and Maximum Monte Carlo (MaxMC) \citep{10.5555/3737916.3738428}. For the sake of simplicity, in the following we refer to Robust PLR as simply PLR.

\subsection{Fine-tuning on Unseen Tasks} 
\label{test tasks}
\begin{table}
   \centering
   \caption{Final win rates after fine-tuning on held-out maps (unseen by all curricula).}
   \label{tab:test tasks performances}
   \begin{center}
   \begin{tabular}{llll}
   \multicolumn{1}{c}{\bf Method}  &\multicolumn{1}{c}{\bf 10m\_vs\_11m}  &\multicolumn{1}{c}{\bf 3s\_vs\_4z}  &\multicolumn{1}{c}{\bf 3s5z}
    \\ \hline \\
    {MAPPO}  & {0.40\tiny{$\pm$0.36}} & {0.04\tiny{$\pm$0.00}} & {0.16\tiny{$\pm$0.36}}\\
    {Random} & {\underline{0.49\tiny{$\pm$0.14}}} & {0.00\tiny{$\pm$0.00}} & {0.59\tiny{$\pm$0.09}}\\
    {Text-Only} & {\underline{0.51\tiny{$\pm$0.18}}} & {0.00\tiny{$\pm$0.00}} & {0.21\tiny{$\pm$0.30}}\\
    {PLR-PVL} & {0.04\tiny{$\pm$0.06}} & {0.00\tiny{$\pm$0.00}} & {0.03\tiny{$\pm$0.05}}\\
    {PLR-MaxMC} & {0.00\tiny{$\pm$0.00}} & {0.00\tiny{$\pm$0.00}} & {0.00\tiny{$\pm$0.00}}\\
    {VIP (ours)} & {\bf{0.53\tiny{$\pm$0.22}}} & {\bf{0.84\tiny{$\pm$0.16}}} & {\bf{0.76\tiny{$\pm$0.20}}}\\
   \end{tabular}
   \end{center}
\end{table}

\begin{figure}
    \centering
    \begin{subfigure}{0.3\textwidth}
        \centering
        \includegraphics[width=1.1\linewidth]{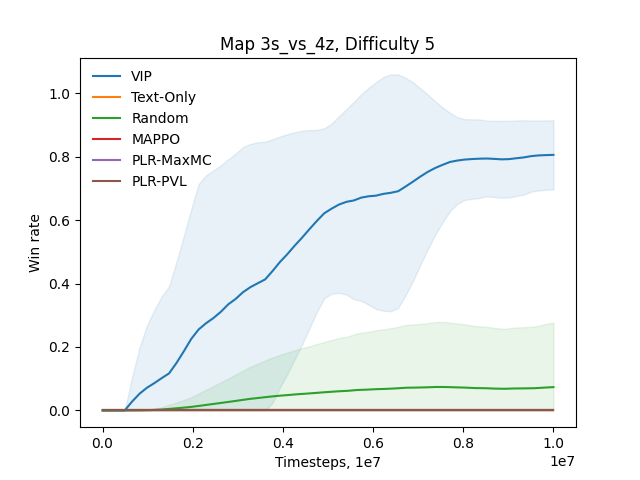}
        \label{fig:ablation 3s_vs_4z scratch}
    \end{subfigure}
    \begin{subfigure}{0.3\textwidth}
        \centering
        \includegraphics[width=1.1\linewidth]{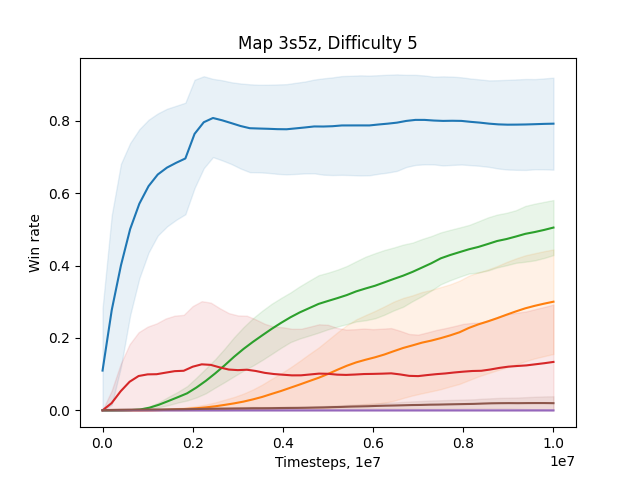}
        \label{fig:ablation 3s5z scratch}
    \end{subfigure}
    \begin{subfigure}{0.3\textwidth}
        \centering
        \includegraphics[width=1.1\linewidth]{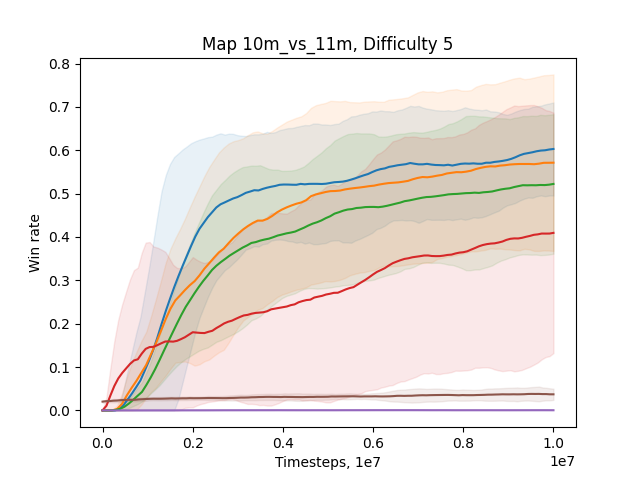}
        \label{fig:ablation 10m_vs_11m scratch}
    \end{subfigure}
    \caption{Fine-tuning win rates on held-out maps (unseen by all curricula).}
    \label{fig:unseen}
\end{figure}

We study generalization to out-of-distribution tasks using the held-out maps: \texttt{10m\_vs\_11m}, \texttt{3s\_vs\_4z}, and \texttt{3s5z}, all at difficulty 5. These maps feature a good variety of team scale and troop types, with two of them also carrying a numbers disadvantage, and are considered among the most challenging in literature \citep{NEURIPS2022_9c1535a0,pmlr-v139-sun21c,Kuba2021TrustRP}. We set the game AI difficulty to 5 since, counterintuitively, it is more challenging than difficulty 7 for MARL (see Appendix \ref{difficulty5} for details). Note that these maps did not appear at any difficulty in any curricula.

Fine-tuning curves are reported in Figure \ref{fig:unseen}, while evaluation win rates after fine-tuning are reported in Table \ref{tab:test tasks performances}. On maps \texttt{3s\_vs\_4z} and \texttt{3s5z}, agents fine-tuned from VIP curricula learn high-performing solutions ($\sim$80\% win rate), while almost all baselines exhibit near-zero win rates. The text-only ablation performs substantially worse than its multi-modal counterpart on these two maps, demonstrating that videos are crucial to the success of VIP and confirming the intuition illustrated in Figure \ref{fig:demo}. On the remaining map \texttt{10m\_vs\_11m}, VIP still obtains the highest performance, comparably with its text-only ablation and random curricula. To better contextualize this result, we note that: (i) \texttt{10m\_vs\_11m} is the only unseen map with marine troops, often considered among the hardest \citep{NEURIPS2022_9c1535a0}; (ii) marine troops appear in $\sim$70\% of tasks from the random curricula and $\sim$15\% of tasks from the text-only ablation curricula, while only appearing once in VIP curricula as the initial task. In all maps, both PLR variants obtain consistently poor performances, showing how scalar scoring can lack essential information when selecting tasks for a curriculum. More detailed analyses of the curricula generated by PLR are provided in Appendix \ref{plr_curricula}.

\subsection{Fine-tuning on Seen Tasks}
\label{seen tasks}
\begin{table}[t]
   \centering
   \caption{Final win rates after fine-tuning on maps frequently seen in almost all curricula; ``$\star$'' entries indicate that the method never chose that map during the curriculum (see Section \ref{sec:seen} for more details).}
    \begin{center}
   \begin{tabular}{llll}
     \multicolumn{1}{c}{\bf Method}  &\multicolumn{1}{c}{\bf 3m}  &\multicolumn{1}{c}{\bf 3s\_vs\_3z}  &\multicolumn{1}{c}{\bf 3s\_vs\_5z}
    \\ \hline \\
     {MAPPO} & {0.97\tiny{$\pm$0.09}} & {0.67\tiny{$\pm$0.06}} & {0.00\tiny{$\pm$0.00}}\\
     {Random} & {\underline{0.99\tiny{$\pm$0.05}}} & {0.42\tiny{$\pm$0.54}} & {0.00\tiny{$\pm$0.00}}\\
     {Text-Only} & {\underline{0.99\tiny{$\pm$0.08}}} & {0.78\tiny{$\pm$0.38}} & {0.00\tiny{$\pm$0.00}}\\
     PLR-PVL & {0.93\tiny{$\pm$0.11}} & {$\star$} & {$\star$}\\
     PLR-MaxMC & {0.90\tiny{$\pm$0.05}} & {0.00\tiny{$\pm$0.00}} & {0.00\tiny{$\pm$0.00}}\\
     {VIP (ours)} & {\bf{1.00\tiny{$\pm$0.00}}} & {\bf{1.00\tiny{$\pm$0.00}}} & {\bf{0.34\tiny{$\pm$0.35}}}\\
   \end{tabular}
   \end{center}
   \label{tab:train tasks performances}
\end{table}
\label{sec:seen}
\begin{figure}
    \centering
    \begin{subfigure}{0.3\textwidth}
        \centering
        \includegraphics[width=1.1\linewidth]{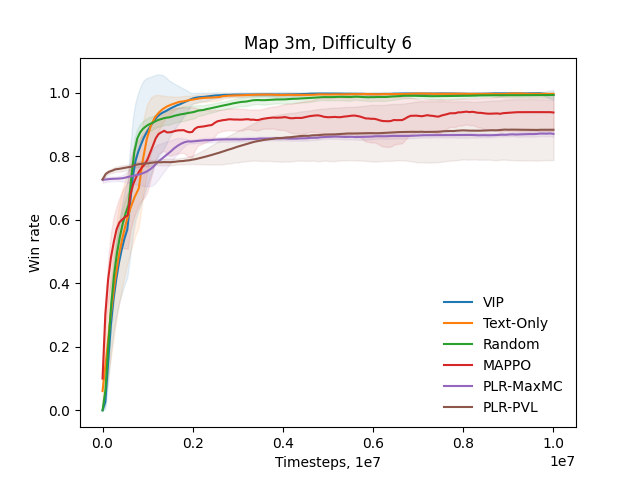}
        \label{fig: 3m difficulty6 forgetting scratch}
    \end{subfigure}
    \begin{subfigure}{0.3\textwidth}
        \centering
        \includegraphics[width=1.1\linewidth]{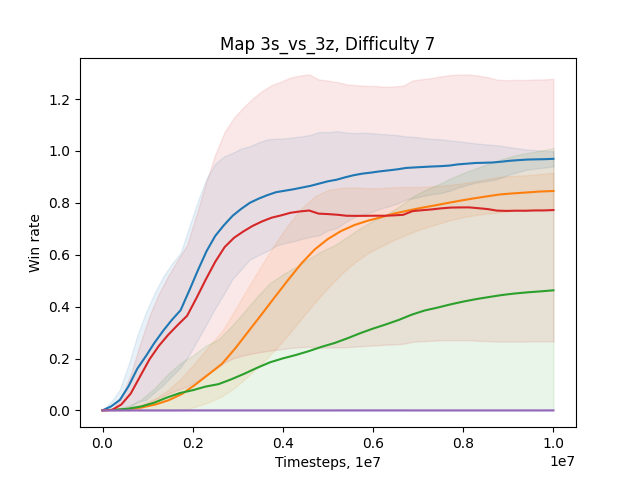}
        \label{fig: 3s_vs_3z difficulty7 forgetting scratch}
    \end{subfigure}
    \begin{subfigure}{0.3\textwidth}
        \centering
        \includegraphics[width=1.1\linewidth]{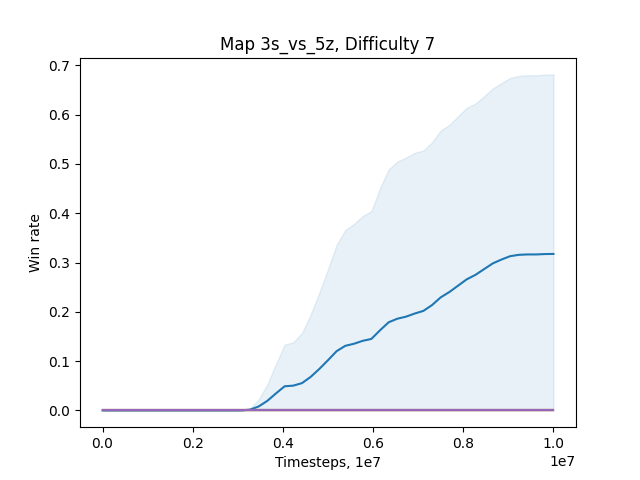}
        \label{fig: 3s_vs_5z difficulty7 forgetting scratch}
    \end{subfigure}
    \caption{Fine-tuning win rates on maps frequently seen in almost all curricula; PLR-PVL is absent where it never chose a specific map during the curriculum (see Section \ref{sec:seen} for more details).}
    \label{fig:seen}
\end{figure}

Alongside the above experiments on generalization to out-of-distribution tasks, we also evaluate fine-tuning performances on 3 maps consistently recommended in almost all curricula. These complement the above results by measuring to what extent each trained policy was able to retain the knowledge used to solve tasks in the early or middle stages of the curriculum. To this end, we selected the following configurations: \texttt{3m} at difficulty 6, \texttt{3s\_vs\_3z} at difficulty 7 and \texttt{3s\_vs\_5z} at difficulty 7. \texttt{3m} is consistently seen in the early stages of all curricula. Maps \texttt{3s\_vs\_3z} and \texttt{3s\_vs\_5z} feature a mismatch in troop types between allies and enemies (paired with a numbers disadvantage in the latter) which significantly increases their complexity \citep{NEURIPS2022_9c1535a0}. The difficulty was chosen as the average among the difficulties chosen by all curricula on each map. We note that the second and third maps were never seen by the PLR-PVL baseline. We still selected these for evaluation since PLR-PVL curricula had no intersections with other curricula outside of \texttt{3m} at difficulty 6 (see Appendix \ref{plr_curricula} for more details). Only strictly selecting tasks seen by all methods, including PLR-PVL, would have thus not yielded a meaningful comparison due to \texttt{3m} not being sufficiently challenging. Even though these same configurations were seen at different stages for different curricula, all curricula included them at the same or later stages than VIP, ensuring no trivial advantage for the latter.

Fine-tuning curves are reported in Figures \ref{fig:seen}, while evaluation win rates after fine-tuning are reported in Table \ref{tab:train tasks performances}. In all maps, VIP still consistently achieves the highest performance. On map \texttt{3s\_vs\_5z}, our method fine-tunes to around 35\% win rate in the allocated number of steps, while all others fail to find any meaningful solution. On map \texttt{3s\_vs\_3z}, VIP shows a significantly better final win rate along with a more stable and faster convergence. All considered methods achieve high win rates on \texttt{3m}. This is expected, since \texttt{3m} is widely recognized as being among the simplest maps in SMAC. Nonetheless, PLR agents show a high starting point in the fine-tuning curve but significantly worse final performances with respect to all other methods. Combined with the low variety observed in PLR curricula (Appendix \ref{plr_curricula}), this again shows that scalar scores can fail to capture learning progress information, resulting in sub-optimal solutions even on easier tasks.

\subsection{Comparison with Grid Search}
\label{grid search}
\begin{table}[t]
   \centering
   \caption{
    Final win rates after fine-tuning on held-out tasks from MAPPO* and VIP. As MAPPO* has access to the best known hyper-parameters for each held-out task found through grid search in \citet{NEURIPS2022_9c1535a0}, it cannot be considered unsupervised.
   }
    \begin{center}
   \begin{tabular}{llllll}
   \multicolumn{1}{c}{\bf Method}  &\multicolumn{1}{c}{\bf 10m\_vs\_11m}  &\multicolumn{1}{c}{\bf 3s\_vs\_4z}  &\multicolumn{1}{c}{\bf 3s5z} &\multicolumn{1}{c}{\bf Env.~interactions} &\multicolumn{1}{c}{\bf Supervised}
    \\ \hline \\
     {MAPPO*} & {\bf{0.84\tiny{$\pm$0.16}}} & {\bf{0.97\tiny{$\pm$0.00}}} & {\underline{0.76\tiny{$\pm$0.26}}} &{$1.7 \times 10^{10}$ (per task)} &Yes\\
     {VIP (ours)} & {0.53\tiny{$\pm$0.22}} & {0.84\tiny{$\pm$0.16}} & {\bf{0.76\tiny{$\pm$0.20}}} &{$1.6 \times 10^8$ (all tasks)} &No\\
   \end{tabular}
   \end{center}
   \label{tab:scratch-dk}
\end{table}

In this section we compare VIP to MAPPO with supervised grid search, which we refer to as MAPPO*, on the held-out maps from Section \ref{test tasks}. MAPPO* trains agents from scratch on each test task using the best known hyper-parameters for each held-out task reported in \citet{NEURIPS2022_9c1535a0}. Since there is no other way to find these values other than brute-force grid search, MAPPO* must assume knowledge of downstream tasks and be able to collect experience on them. Because testing a single hyper-parameters combination requires the same steps as each task in VIP, we can directly compare the number of combinations searched in \citet{NEURIPS2022_9c1535a0} with the length of our curricula. Hyper-parameters for MAPPO* were found by searching around 1700 combinations\footnote{Table 13 from \citet{NEURIPS2022_9c1535a0} implies $3^3 \cdot 2^6 = 1728$ combinations were searched, with each combination taking $10^7$ environment interactions.}, while our curricula only contain 16 tasks and thus use around 100 times less steps. Moreover, values found via grid search are only optimal for a single task, meaning that the number of environment steps needed for $N$ test tasks further scales with $N$. Any open-ended curriculum method, including VIP and PLR, instead only requires the same number of steps regardless of the number of test tasks. Nonetheless, comparisons with brute-force supervised search still help contextualize these methods.

Final performances are reported in Table \ref{tab:scratch-dk}; fine-tuning curves are reported in Appendix \ref{dk}. As expected, the considerably higher number of supervised environment interactions results in MAPPO* outperforming VIP on two out of three maps. On the remaining map, VIP obtains comparable performances with MAPPO* but shows faster convergence.

While results obtained through extensive grid search are relevant, we argue that developing methodologies that acquire domain knowledge in more intelligent ways with significantly fewer interactions is just as important for the development of open-ended and lifelong-learning agents. The impact of such methodologies does not necessarily lie in an immediate advancement of the state-of-the-art performance on a specific benchmark, but also in: (i) being reusable in more complex systems or any setting where grid searches of such scale are inaccessible; (ii) making results more reproducible by the scientific community of RL \citep{Henderson_Islam_Bachman_Pineau_Precup_Meger_2018}.

\section{Conclusions}
We introduced Visual Inspection of Policies (VIP), a new method to automatically generate open-ended curricula for Reinforcement Learning (RL) agents. VIP uses a VLM to directly inspect episode videos from the current policy and recommend tasks tailored to the current learning progress. It is also independent of the number of controllable agents in the environment as long as they are rendered in a single video. We have studied the effectiveness of VIP empirically with a lightweight and openly accessible VLM on a complex, strategic multi-agent domain, showing that policy videos play a key role in VIP's performance through its text-only ablation. At the same time, our results show that VIP outperforms autocurricula which rely on scalar task scores.

Future work will investigate the use of other modalities and multi-modal LLMs for unsupervised autocurricula, possibly combining more than one LLM such as in more complex text-only approaches \citep{ICLR2024_18b40f12,ICLR2025_d40d7cbe}. These developments are likely to require more advanced sanitization techniques, especially when considering larger or infinite task spaces. More in general, we envision VIP as only the tip of the iceberg in leveraging VLMs and other multi-modal LLMs as a natural avenue for open-ended Reinforcement Learning: much like text is only one of the modalities through which humans guide their unsupervised learning processes, rich visual and multi-modal information is also essential when evaluating the progress of any artificial learning agent.

\bibliography{collas2026_conference}
\bibliographystyle{collas2026_conference}

\newpage
\appendix

\section{Experimental Details}
\label{experimental details}
Here we report additional experimental details on every aspect outside of the VLM. These are instead reported in Section \ref{vlm details}.

\subsection{Train-Test Map Subdivision}
\begin{table}[t]
  \centering
  \caption{Train-test map subdivision of VIP. 12 train maps and 3 test maps correspond to a 80\%-20\% split of the original set of maps available in SMAC.}
  \label{tab:train test maps}
  \begin{center}
  \begin{tabular} {ll}
  \multicolumn{1}{c}{\bf Train (12)}  &\multicolumn{1}{c}{\bf Test (3)}
    \\ \hline \\
  {2m\_vs\_1z} & {10m\_vs\_11m}\\
  {2s3z} & {3s\_vs\_4z}\\
  {3m} & {3s5z}\\
  {3s5z\_vs\_3s6z} & {}\\
  {3s\_vs\_3z} & {}\\
  {3s\_vs\_5z} & {}\\
  {5m\_vs\_6m} & {}\\
  {6h\_vs\_8z} & {}\\
  {8m} & {}\\
  {8m\_vs\_9m} & {}\\
  {25m} & {}\\
  {27m\_vs\_30m} & {}\\
  \end{tabular}
  \end{center}
\end{table}

Table \ref{tab:train test maps} contains the train-test subdivision of SMAC maps for VIP. This subdivision is motivated in Section \ref{test tasks}.

\subsection{Hyper-Parameters}
\begin{table}[t]
  \centering
  \caption{Hyper-parameters used for all experiments outside of Sections \ref{grid search} and \ref{dk}. They correspond to the default set in MARLlib \citep{hu2022marllib}.}
  \label{tab:hyper-parameters}
  \begin{center}
  \begin{tabular} {ll}
    \multicolumn{1}{c}{\bf Hyper-parameter}  &\multicolumn{1}{c}{\bf Value}
    \\ \hline \\
  {timesteps} & {1e7}\\
  {optimizer} & {Adam}\\
  {core\_arch} & {gru}\\
  {fc\_layer} & {1}\\
  {hidden\_state\_size} & {256}\\
  {out\_dim\_fc} & {128}\\
  {share policy} & {all}\\
  {lr} & {0.0005}\\
  {framework} & {torch}\\
  {num\_workers} & {32}\\
  {num\_gpus} & {1}\\
  {batch size} & {3200}\\
  {lambda} & {1.0}\\
  {kl coefficient} & {0.2}\\
  {batch episode} & {10}\\
  \end{tabular}
  \end{center}
\end{table}

Table \ref{tab:hyper-parameters} reports the hyper-parameters used for all experiments outside of Sections \ref{grid search} and \ref{dk}. The choice of a GRU-based architecture makes sense for SMAC due to the severe partial observability of the environment. We refer to \citet{hu2022marllib} for more details on model architectures.

\subsection{SMAC Details}
\label{SMAC details}
\begin{figure}
  \centering
  \includegraphics[width=.30\textwidth]{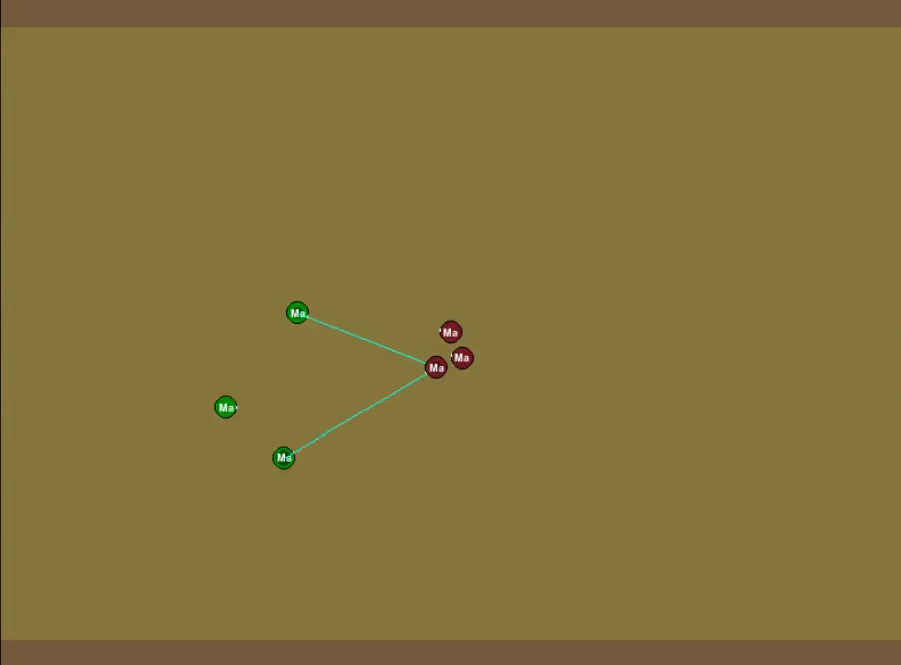}
  \caption{A frame from the rendering of the \texttt{3m} map from SMAC. This scheme makes it often unclear which units are shooting and which are being shot, since the color of the shooting lines is the same for both teams.}
  \label{fig:smac render}
\end{figure}

SMAC is a partially observable environment, with each agent retrieving local observations from their own field of view. Local observations contain, for each ally and enemy troop within sight: \emph{Euclidean distance}, \emph{distance along x}, \emph{distance along y}, \emph{health}, \emph{shield}, \emph{unit typology}. The full state of the environment includes the above information for all entities in the map, without any sight restrictions. The action space is discrete, with available actions being: \emph{move north}, \emph{move east}, \emph{move west}, \emph{move south}, \emph{attack}$_i$ (one for each enemy $i$), \emph{stop}, \emph{no-op}.

The built-in rendering scheme for SMAC is minimal, which has both positive and negative effects. On the one hand, it does not employ purely cosmetic representations, which makes it free of distractors. At the same time, it might also lose information in representing the state of the game (see Figure \ref{fig:smac render}).

Since the size and semantics of the observations and actions depend on the number and types of troops, transferring policies between maps is not possible in vanilla SMAC. To this end, we have padded the observations and action spaces of all maps to a unique size, corresponding to the map with the highest number of agents (\texttt{27m\_vs\_30m}). The padding simply consisted of zeros which have no effects on environment state transitions. Even though this is the simplest solution to the size mismatch between maps, it still adds distractors to each task, which can lower the sample efficiency of MARL training.

\section{Difficulty 5 is More Challenging than 7}
\label{difficulty5}
\begin{figure*}[t]
    \centering
    \begin{subfigure}{0.3\textwidth}
        \centering
        \includegraphics[width=1.1\linewidth]{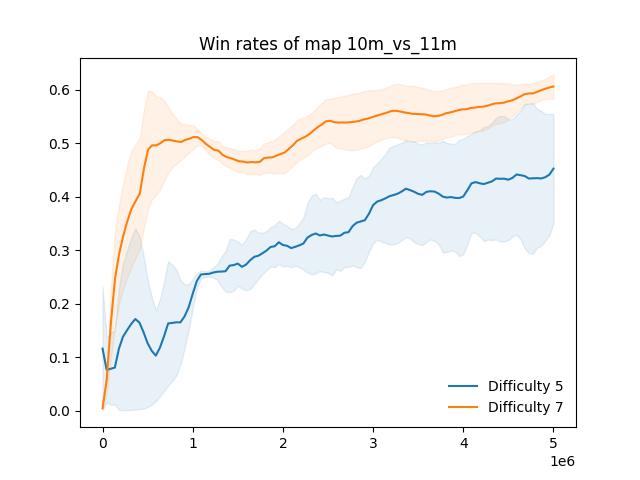}
        \label{fig: 10m_vs_11m difficulty5-7}
    \end{subfigure}
    \begin{subfigure}{0.3\textwidth}
        \centering
        \includegraphics[width=1.1\linewidth]{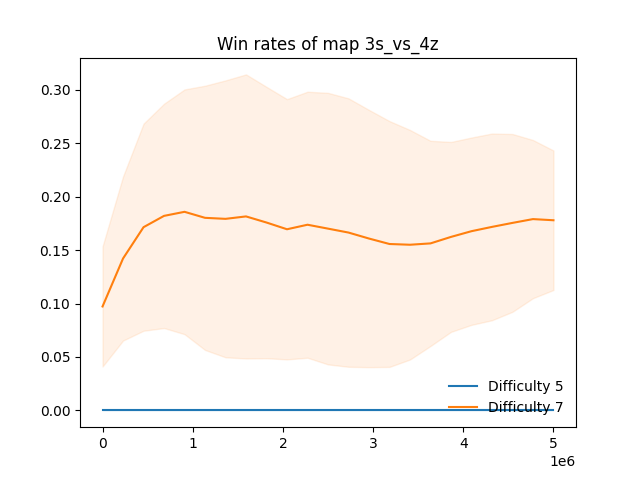}
        \label{fig: 3s_vs_4z difficulty5-7}
    \end{subfigure}
    \begin{subfigure}{0.3\textwidth}
        \centering
        \includegraphics[width=1.1\linewidth]{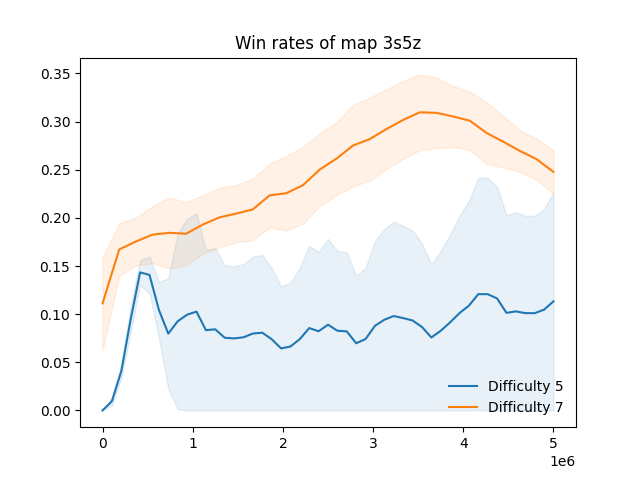}
        \label{fig: 3s5z difficulty5-7}
    \end{subfigure}
    \caption{Training curves of vanilla MAPPO over the three test maps with difficulties 5 and 7, averaged over 5 seeds with shaded areas corresponding to standard deviation. Difficulty 5 displays significantly worse convergence, which counterintuitively makes it harder than difficulty 7 for MARL agents.}
    \label{fig:diff5vs7}
\end{figure*}

Existing literature usually sets the difficulty of most SMAC maps to 7 \citep{NEURIPS2022_9c1535a0,PORTAL}. Here we motivate our choice of instead choosing difficulty 5 for our main experiments on unseen maps (Section \ref{test tasks}): while experimenting with the game AI difficulty setting, we found that difficulty 5 poses a harder challenge to MARL agents than difficulty 7. To confirm this, we trained MAPPO from scratch on both difficulties on the maps used as unseen maps in our main experiments. Results are shown in Figure \ref{fig:diff5vs7} and confirm that difficulty 5 is significantly harder on these maps for MARL agents. While this may seem counterintuitive at first, we note that the SMAC game AI difficulty levels were originally developed for human players rather than RL agents. This might make it possible that some strategies of the game AI at difficulty 5 are easier to beat for human players, but may be much harder for MARL. Vice-versa, the game AI might play out better against humans rather than MARL agents on difficulty 7.

\section{Plots for Comparison with grid search}
\label{dk}

\begin{figure*}[t]
    \centering
    \begin{subfigure}{0.3\textwidth}
        \centering
        \includegraphics[width=1.1\linewidth]{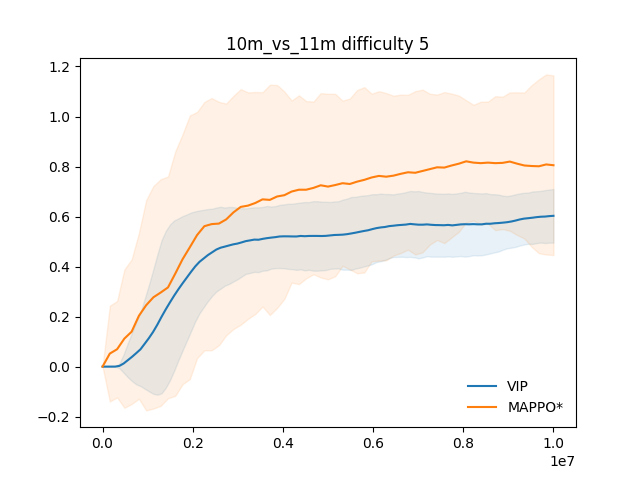}
        \label{fig: 10m_vs_11m scratch dk}
    \end{subfigure}
    \begin{subfigure}{0.3\textwidth}
        \centering
        \includegraphics[width=1.1\linewidth]{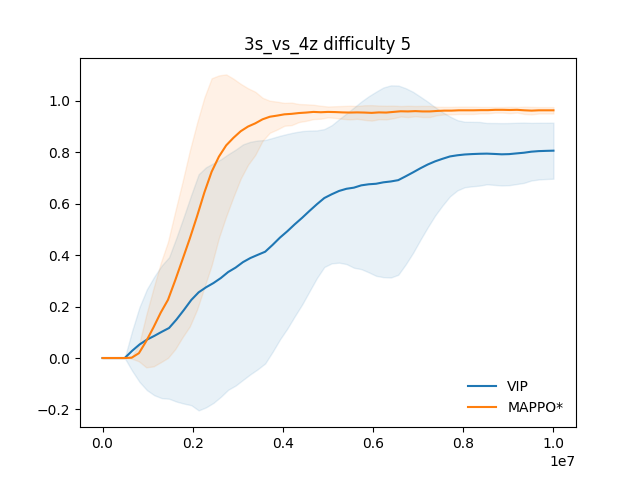}
        \label{fig: 3s_vs_4z scratch dk}
    \end{subfigure}
    \begin{subfigure}{0.3\textwidth}
        \centering
        \includegraphics[width=1.1\linewidth]{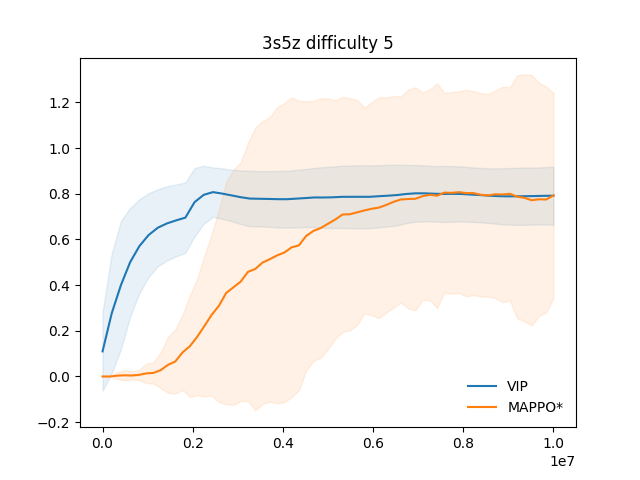}
        \label{fig: 3s5z scratch dk}
    \end{subfigure}
    \caption{Fine-tuning win rates for unseen tasks, where shaded areas represent one standard deviation over 5 seeds, for comparison of VIP against MAPPO with extensive grid search (MAPPO*). See Section \ref{grid search} for more details.}
    \label{fig:mappostar}
\end{figure*}

Figure \ref{fig:mappostar} reports the fine-tuning curves for the experiment described in Section \ref{grid search}. MAPPO* shows better convergence than VIP on 2 out of the 3 unseen maps. On map \texttt{3s5z} VIP instead achieves a similar win rate to MAPPO* with significantly faster convergence. We refer to Section \ref{grid search} for a more thorough discussion of these results.

\section{Additional Qualitative Analyses}
\label{sec:qualitative}
\begin{figure}
    \centering
    \includegraphics[width=1.0\linewidth]{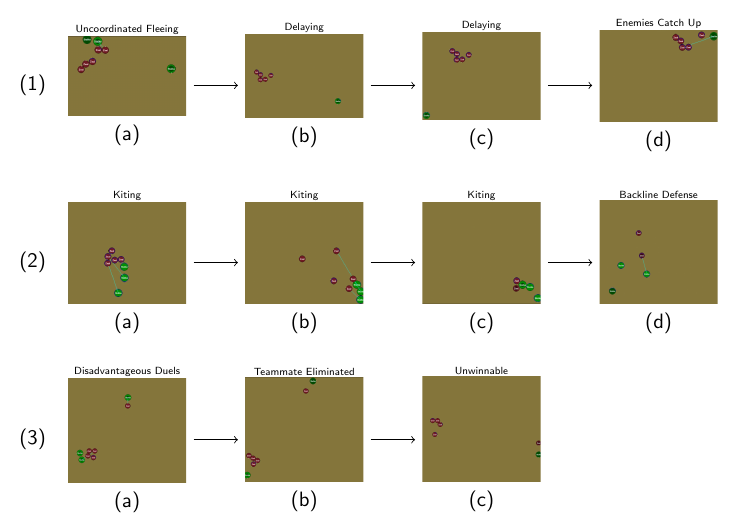}
    \caption{(1) Negative example from a text-only ablation curriculum that recommended map \texttt{3s\_vs\_5z}, which features numbers disadvantage, after finding success on the previous map \texttt{3s\_vs\_3z}, which is instead balanced in numbers. While the simple strategy of fleeing from enemies was moderately successful with balanced numbers, it completely fails when allies are at a numbers disadvantage and results in the chosen task being too difficult to make any progress. (2) A positive example from VIP where it recommended the same map of the previous example (\texttt{3s\_vs\_5z}). Contrary to the previous example, here agents are visibly able to \emph{kite}: skillfully alternating movement and attacks, without letting enemies get close --- in frame (d), the agents are even able to keep formation by moving the most damaged agent (dark green) to the back of the formation. Despite the $\sim$10\% win rate due to timeouts, VIP recognizes the behavior and recommends the same task, resulting in around 50\% win rate, much higher than what the text-only ablation obtains in examples (1) and (3) in a similar scenario. (3) A continuation of example (1) from the same curriculum. The text-only ablation again recommended map \texttt{3s\_vs\_5z}, in which simple fleeing strategies fail. From this point on, the text-only ablation continues to alternate the recommendation of \texttt{3s\_vs\_5z} with easier maps in which fleeing can be moderately successful, without being able to conclude the main reason behind the agent's failure. This results in oscillating win rates that, however, did not advance the capabilities of the agents, which continue to employ a basic fleeing strategy without the coordination needed for proper kiting.}
    \label{fig:qualitative}
\end{figure}

Figure \ref{fig:qualitative} reports additional case studies of curriculum recommendations in which episode videos contain crucial information that might be absent from local learning signals. These include both positive examples (beneficial task recommendation by VIP), and negative examples (inconclusive recommendations by the text-only ablation).




\section{Compute Resources}
\label{compute resources}
All our experiments used a single Tesla V100-SXM2-32GB. This GPU features 32 gigabytes of VRAM. The storage footprint of experimental data consists in approximately 20 gigabytes. Each training segment of 1e7 timesteps lasted around 6 hours when using MARLlib \citep{hu2022marllib} and around 18 hours when using the original MAPPO implementation \citep{NEURIPS2022_9c1535a0}. Each VLM inference lasted approximately 4 minutes, which results in the VLM inference contributing a negligible amount (around 1\%) to the compute time spent for each VLM-based curriculum.

\section{VLM Details}
\label{vlm details}
Here we provide additional experimental details on the VLM, such as prompt templates, example VLM outputs and quantitative analysis of VLM failure modes. All our experiments used the 7-billion version of VideoLLaMa2.

\subsection{VLM Hyper-parameters}
We use the following hyper-parameters for this VLM: \emph{temperature} of 0.4, \emph{top\_p} of 0.8. Both were set in preliminary experiments and are aimed at balancing the probability of hallucinations and the variety of the task recommendations.

\subsection{VLM Output Sanitization and Failure Modes}
\label{sec:xp:sanitize}
Here we quantify the failure modes of the VLM as a task selector along with the sentence similarity module. The VLM hallucinated exclusively once per curriculum, resulting in a moderate 5.6\% rate. By inspecting these cases, we confirmed that the sentence similarity module correctly handled the entirety of the hallucinations, meaning no other fallback mechanism was triggered in practice.

\subsection{Prompt Templates}
\label{sec:prompt-templates}
We follow the standard practice of providing the VLM with an initial \emph{system prompt}, followed by multiple prompts, each asking for a task recommendation for the curriculum. The system prompt assigns the VLM the role of a ``video game expert'', whose duty consists of evaluating the performances of RL agents by means of inspecting policy videos and other information. The system prompt also includes information on the domain such as a small description of the SMAC environment, the list of difficulties, train maps, etc. The system prompt also specifies the desired format for the task recommendation output. At each curriculum step, the task prompt reiterates a summary of the task assigned to the VLM and asks for a task recommendation based on the provided visual and other information. Prompt templates for the system prompt and task prompts are reported below.

The inclusion of repeated contextual information, such as the list of previously selected tasks, was needed in our experiments due to limited context capabilities of the 7B model. We confirmed that this was the case by simulating interactions with different levels of context repetition in task prompts. Specifically, we compared: (i) no context information; (ii) previous step context (last selected task); (iii) complete context information (full list of previous tasks). We report that the cases with limited (ii) or no context information (i) showed a lack of awareness of past interactions by repeatedly selecting tasks that were selected in the previous few steps. Instead, when provided with complete contextual information (iii), the 7B model properly accounted for it as expected, motivating the final task prompt template shown below.

Each textual prompt during the curricula was paired with a single video 1024x1024 at 25 frames per second; while video length varied based on the episode length, in our experiments all videos were 1-10 seconds long, since the maximum episode timeout for a SMAC map is 250 steps.

\subsubsection{System Prompt Template}
\begin{quote}
    You are a video game expert in charge of evaluating the performances of a reinforcement learning algorithm through the visualization of recorded videos of the algorithm's actions within a StarcraftII environment.
    You must evaluate its performance on that specific setting, i.e. on that specific map, and its potential ability to generalize to other maps of the same environment, i.e. game.
    Finally, you must select the next 'interesting', reachable task.
    
    The environment is a simplified version of the StarcraftII game: two adversarial teams fight, the first team losing all its components loses the battle. StarcraftII is a strategic, military-inspired, game where attacking takes the form of shooting or close combat, depending on the typology of troops deployed.
    Its reinforcement learning environment is multi-agent, specifically the algorithm commands all the agents of a single team, thereafter restricting the setting to exclusively a cooperative one.
    The videos of this environment that will be shown to you depict circles of two colours: on one side the green circles compose the allies team, thereafter the one commanded by the reinforcement learning agents, on the other side the red circles compose the enemies team.
    
    The list of the selectable maps is the following:
    \begin{itemize}
        \item 2 Marines vs 1 Zealot,
        \item 2 Stalkers \& 3 Zealots vs 2 Stalkers \& 3 Zealots,
        \item 3 Marines vs 3 Marines,
        \item 3 Stalkers \& 5 Zealots vs 3 Stalkers \& 6 Zealots,
        \item 3 Stalkers vs 3 Zealots,
        \item 3 Stalkers vs 5 Zealots,
        \item 5 Marines vs 6 Marines,
        \item 6 Hydralisks vs 8 Zealots,
        \item 8 Marines vs 8 Marines,
        \item 8 Marines vs 9 Marines,
        \item 25 Marines vs 25 Marines,
        \item 27 Marines vs 30 Marines
    \end{itemize}    
    
    The map names represent the typologies, and number, of troops deployed by the two teams, the name before the 'vs' represents the team of the reinforcement learning agents, i.e. the allies team, while the name after the 'vs' represents the enemy team.
    The four races, i.e. typologies, of troops are : Stalkers, Zealots, Marines, Hydralisks.
    Within the videos which will be shown you, the troop typology is visible because an abbreviation of their typology is written on the circle representing the troop. The abbreviations are: 'Zeal' for Zealots, 'Hydra' for Hydralisks, 'Ma' for Marines, 'Stalke' for Stalkers.
    For instance, a Zealot allied troop will be shown on the map by a green circle with 'Zeal' written on it.
    As another instance, a Marine enemy troop will be shown on the map by a red circle with 'Ma' written on it.
    
    Zealots can only attack melee, they are strong against Hydralisks. Marines have ranged attacks, they possess high HP, granting them more survivability. Stalkers are ranged attackers. Hydralisks have both melee and ranged attacks, but they are relatively fragile.
    
    There are ten levels of difficulties, from 1 to 10, it follows the full description of the levels of difficulties by the game environment creators:
    difficulties = \{
        '1': 'Very Easy',
        '2': 'Easy',
        '3': 'Medium',
        '4': 'Medium Hard',
        '5': 'Hard',
        '6': 'Harder',
        '7': 'Very Hard',
        '8': 'Super Hard',
        '9': 'Ultra Hard',
        '10': 'Max'
    \}
    
    The maps and difficulties can form any combination, select the map and the difficulty independently, nevertheless pick them with a clear idea of the combination you wish to obtain.
    For example if you want to try map '8 Marines vs 9 Marines' with and a grade of difficulty of '5', you will obtain the combination '8 Marines vs 9 Marines with difficulty Hard'.
    You shall select the most appropriate combination from the parameters which are instructed you below.
    
    Do not repeat on a combination of map and difficulty present within the history of tasks: if the win rate is high (e.g. win rate \begin{math}>\end{math} 70\%), increase the difficulty level or turn your attention to another map, if the win rate is too low (e.g. win rate \begin{math}<\end{math} 30\%), decrease the difficulty level or turn your attention to another map.
    Your evaluation will bear the objective of suggesting the next task, which must be 'interesting' for the reinforcement learning algorithm, i.e. avoid tasks which are too redundant, nevertheless it must be not too difficult for the current capabilities of the agents.
    For example, transitioning from task '3 Marines vs 3 Marines with difficulty Very Easy', to task '3 Stalkers vs 5 Zealots with difficulty Ultra Hard' is a humongous leap which is likely not within the current capabilities of the algorithm . On the flip side, transitioning from task '3 Marines vs 3 Marines with difficulty Very Easy' to task '3 Marines vs 3 Marines with difficulty Easy', is a step the algorithm is likely capable of making, nevertheless the two tasks are very redundant, thereafter labeling the selected task 'not interesting'.
\end{quote}

\subsubsection{Task Prompt Template}
\begin{quote}
    The video shows the current capability of the agents on task 'task\_name', which is the task you previously selected. The average win rate of the agents on the current task is average\_test\_win\_rate.
    Recall that the green circles are the allies, while the red circles are the enemies.
    What is the next task, not too advanced for the agents current capabilities, which you judge 'interesting', among the possible ones?
    Remember to clearly state the task name in the format provided you earlier, strictly selecting a combination of the list of maps and difficulties listed within the system prompt.
    History of tasks you have selected:
    Do not repeat combinations of map and difficulty present within the history of tasks: if the win rate of the current task is high ( e.g. win rate \begin{math}>\end{math} 70\%), for the next task increase the difficulty level or turn your attention to another map, if the win rate of the current task is too low (e.g. win rate \begin{math}<\end{math} 30\%), for the next task decrease the difficulty level or turn your attention to another map.
\end{quote}

\subsection{Example VLM Outputs}
We report three examples of VLM outputs from VIP runs below. These examples show the observed variety in answer length and complexity, and also motivate the need for the sentence similarity module even when the VLM does not hallucinate task specifications.

\subsubsection{First Example}
\begin{quote}
    Based on the information provided, I will select a new task that is not too advanced for the agents' current capabilities and is considered 'interesting'. 
    Since the win rate of the current task (3 Marines vs 3 Marines with difficulty extreme) is high (96\%), I will increase the difficulty level or switch to another map. However, since the difficulty 'max' is already the highest, I will switch to another map.
    
    I propose the following combination:
    **Map:** 8 Marines vs 9 Marines
    **Difficulty:** Medium Hard (4)
    
    This combination is chosen because it represents a moderate challenge for the algorithm, with a balanced number of troops on each side. The difficulty level of "Medium Hard" provides a reasonable challenge without being overly difficult, allowing the algorithm to demonstrate its capabilities while still being able to learn and adapt.
    The win rate for this combination is expected to be around 50-60\%, which is neither too high nor too low. This will allow the algorithm to continue learning and improving, while also providing a clear indication of its performance.
    By selecting this combination, we can evaluate the algorithm's ability to adapt to a slightly more challenging environment, while still maintaining a balance between difficulty and redundancy. This will help us identify areas where the algorithm can improve, and provide a solid foundation for future tasks.
\end{quote}

\subsubsection{Second Example}
\begin{quote}
    Based on the performance of the agents and the history of tasks, I will suggest a new task that is both challenging and interesting for the algorithm.
    I propose the following task: 3 Stalkers \& 5 Zealots vs 3 Stalkers \& 5 Zealots with difficulty Max.
    This task is not too advanced for the agents' current capabilities, as they have already demonstrated high win rates on similar tasks. The combination of map and difficulty level provides a good challenge for the algorithm to learn from and improve upon.
\end{quote}

\subsubsection{Third Example}
\begin{quote}
    Based on the current capabilities of the agents and the history of tasks, I would suggest selecting the task "10 Marines vs 11 Marines" with difficulty level "7" (Very Hard). This task is not too advanced for the agents' current capabilities and provides a new challenge for them to learn and adapt to. Additionally, this task has not been previously selected in the history of tasks, ensuring that the agents are not simply repeating the same task with different parameters.
\end{quote}

\section{PLR Curricula Analysis}
\label{plr_curricula}
Here we provide additional details on the poor performances of PLR curricula observed in the main experiments. To showcase differences with VIP-generated curricula, below we report one example curriculum for each PLR baseline and one example curriculum from VIP, with each line containing the chosen task as the pair (map, difficulty). We also report the task distributions of these examples in Table \ref{tab:currs}. While the distributions highlight the low variety of curricula from PLR (especially with the PVL score), the examples theselves show the PLR curricula to be much more erratic in the choice of both maps and difficulty: several repeated tasks are often not adjacent in the curriculum. On the other hand, VIP generates a more linear progression both in maps and difficulty, often repeating the same map with increasing difficulty and almost never going back to a task in non-adjacent curriculum steps. We hypothesize the poorly-generated curricula from PLR are due to a mixture of: (i) lack of information present in scalar score; and (ii) scalar score themselves being known to be vulnerable to task that are too hard \citep{10.5555/3737916.3738428}.

\begin{table}[t]
  \centering
  \caption{Task Distributions for the example curricula in Section \ref{plr_curricula}.}
  \label{tab:currs}
  \begin{center}
  \begin{tabular} {llr}
  \multicolumn{1}{c}{\bf Method}  &\multicolumn{1}{c}{\bf Task}  &\multicolumn{1}{c}{\bf Occurrences}
    \\ \hline \\
  \multirow{5}{*}{PLR-PVL} 
&    ('3m', '6') & 1 \\
&    ('2m\_vs\_1z', '8') & 14 \\
&    ('3m', '8') & 2 \\
&    ('5m\_vs\_6m', '9') & 1 \\
&    ('8m\_vs\_9m', '9') & 1 \\
  \midrule
  \multirow{10}{*}{PLR-MaxMC} 
&  ('3m', '6') & 2 \\
&  ('25m', '3') & 1 \\
&  ('3s\_vs\_5z', '3') & 5 \\
&  ('2m\_vs\_1z', '3') & 4 \\
&  ('3m', '8') & 1 \\
&  ('5m\_vs\_6m', '5') & 2 \\
&  ('3m', '5') & 1 \\
&  ('3s\_vs\_5z', '7') & 1 \\
&  ('3s\_vs\_3z', '6') & 1 \\
&  ('3s\_vs\_5z', '8') & 2 \\
  \midrule
  \multirow{12}{*}{VIP}
&  ('3m','6') & 2 \\
&  ('3s\_vs\_3z','4') & 1 \\
&  ('3s\_vs\_3z','5') & 1 \\
&  ('3s\_vs\_3z','7') & 2 \\
&  ('3s\_vs\_3z','8') & 1 \\
&  ('3s\_vs\_5z','6') & 1 \\
&  ('3s\_vs\_5z','7') & 2 \\
&  ('3s\_vs\_5z','8') & 1 \\
&  ('3s5z\_vs\_3s6z','4') & 3 \\
&  ('3s5z\_vs\_3s6z','5') & 1 \\
&  ('3s5z\_vs\_3s6z','8') & 1 \\
&  ('3s5z\_vs\_3s6z','7') & 2 \\
  \end{tabular}
  \end{center}
\end{table}

\subsection{Example from PLR-PVL}
\begin{quote}
    ('3m', '6') \\
    ('2m\_vs\_1z', '8') \\
    ('3m', '8') \\
    ('2m\_vs\_1z', '8') \\
    ('2m\_vs\_1z', '8') \\
    ('2m\_vs\_1z', '8') \\
    ('2m\_vs\_1z', '8') \\
    ('2m\_vs\_1z', '8') \\
    ('2m\_vs\_1z', '8') \\
    ('2m\_vs\_1z', '8') \\
    ('3m', '8') \\
    ('5m\_vs\_6m', '9') \\
    ('8m\_vs\_9m', '9') \\
    ('2m\_vs\_1z', '8') \\
    ('2m\_vs\_1z', '8') \\
    ('2m\_vs\_1z', '8') \\
    ('2m\_vs\_1z', '8') \\
    ('2m\_vs\_1z', '8') \\
    ('2m\_vs\_1z', '8') \\
\end{quote}

\subsection{Example from PLR-MaxMC}
\begin{quote}
    ('3m', '6') \\
    ('25m', '3') \\
    ('3m', '6') \\
    ('3s\_vs\_5z', '3') \\
    ('3s\_vs\_5z', '3') \\
    ('3s\_vs\_5z', '3') \\
    ('3s\_vs\_5z', '3') \\
    ('2m\_vs\_1z', '3') \\
    ('3m', '8') \\
    ('5m\_vs\_6m', '5') \\
    ('2m\_vs\_1z', '3') \\
    ('3s\_vs\_5z', '3') \\
    ('3m', '5') \\
    ('3s\_vs\_5z', '7') \\
    ('5m\_vs\_6m', '5') \\
    ('3s\_vs\_3z', '6') \\
    ('3s\_vs\_5z', '8') \\
    ('3s\_vs\_5z', '8') \\
    ('2m\_vs\_1z', '3') \\
    ('2m\_vs\_1z', '3') \\
\end{quote}

\subsection{Example from VIP}
\begin{quote}
    ('3m','6') \\
    ('3m','6') \\
    ('3s\_vs\_3z','4') \\
    ('3s\_vs\_3z','5') \\
    ('3s\_vs\_3z','7') \\
    ('3s\_vs\_3z','7') \\
    ('3s\_vs\_3z','8') \\
    ('3s\_vs\_5z','6') \\
    ('3s\_vs\_5z','7') \\ 
    ('3s\_vs\_5z','7') \\
    ('3s\_vs\_5z','8') \\
    ('3s5z\_vs\_3s6z','4') \\
    ('3s5z\_vs\_3s6z','4') \\
    ('3s5z\_vs\_3s6z','4') \\
    ('3s5z\_vs\_3s6z','5') \\
    ('3s5z\_vs\_3s6z','8') \\
    ('3s5z\_vs\_3s6z','7') \\
    ('3s5z\_vs\_3s6z','7') \\
\end{quote}

\end{document}